# Using exoskeletons to assist medical staff during prone positioning of mechanically ventilated COVID-19 patients: a pilot study

Serena Ivaldi[1], Pauline Maurice[1], Waldez Gomes[1], Jean Theurel[2], Lien Wioland[2], Jean-Jacques Atain-Kouadio[2], Laurent Claudon[2], Hind Hani[3], Antoine Kimmoun[4], Jean-Marc Sellal[5], Bruno Levy[4], Jean Paysant[6], Serguei Malikov[7], Bruno Chenuel, Nicla Settembre[7,3]

[1] University of Lorraine, CNRS, Inria, LORIA, F-54000 Nancy, France
{serena.ivaldi, waldez.azevedo-gomes-junior}@inria.fr, pauline.maurice@loria.fr
[2] Working Life Department, French National Research and Safety Institute for the Prevention of Occupational Accidents and Diseases (INRS), F-54500 Vandoeuvre-les-Nancy, France
{jean.theurel, lien.wioland, jean-jacques.atain-kouadio, laurent.claudon}@inrs.fr
[3] Virtual Hospital of Lorraine, CUESim, University of Lorraine, F-54000 Nancy, France
hind.hani@univ-lorraine.fr
[4] Medical Intensive Care Unit Brabois, Nancy University Hospital, INSERM U1116, University of Lorraine, 54000 Nancy, France
{a.kimmoun, b.levy}@chru-nancy.fr
[5] CHRU-Nancy, Department of Cardiology, F-54000 Nancy, France
jm.sellal@chru-nancy.fr
[6] CHRU-Nancy, Department of Rehabilitation Medicine, EA DevAH, University of Lorraine, F-54000 Nancy, France
jean.paysant@univ-lorraine.fr
[7] CHRU-Nancy, Inserm 1116, University of Lorraine, F-54000 Nancy, France
{s.malikov,n.settembre}@chru-nancy.fr
[8] CHRU-Nancy, University of Lorraine, University Centre of Sports Medicine and Adapted Physical Activity, Pulmonary Function and Exercise Testing Department, EA DevAH, Department of Medical Physiology, F-54000 Nancy, France
b.chenuel@chru-nancy.fr

**Abstract.** We conducted a pilot study to evaluate the potential and feasibility of back-support exoskeletons to help the caregivers in the Intensive Care Unit (ICU) of the University Hospital of Nancy (France) executing Prone Positioning (PP) maneuvers on patients suffering from severe COVID-19-related Acute Respiratory Distress Syndrome. After comparing four commercial exoskeletons, the Laevo passive exoskeleton was selected and used in the ICU in April 2020. The first volunteers using the Laevo reported very positive feedback and reduction of effort, confirmed by EMG and ECG analysis. Laevo has been since used to physically assist during PP in the ICU of the Hospital of Nancy, following the recrudescence of COVID-19, with an overall positive feedback.

## 1   Introduction

The COVID-19 pandemic has stressed healthcare systems worldwide as never before, with significant consequences for clinical management, including rationing of care, and facing a limitation of capacity and resources of Intensive Care Units (ICU) to safely maintain a high number of patients on mechanical ventilation during the surge. Prone positioning (PP), i.e. when a patient is repositioned from a supine position (i.e., lying on his/her back) to lie on a prone position (i.e., on his/her front side), is known to improve oxygenation and ventilatory mechanics in Acute Respiratory Distress Syndrome (ARDS) patients who require mechanical ventilatory support [1]. PP is therefore largely used during the COVID-19 pandemic. For instance, in the Hospital of Nancy, the ICU performed 116 PP maneuvers in the first 10 days of the outbreak, which is equivalent to the number of maneuvers they usually perform in a whole year. Although turning a patient into the prone position is not an invasive procedure, it is complex and has many potential complications requiring adequate and well-trained healthcare staff (Fig. 1). It is also an exhausting and very time-consuming task. Namely, PP procedure requires the medical staff to remain with their torso bent forward for several minutes, thus causing load and potential injuries in the low back [2]. In addition, obesity-related complications have been identified as risk factors of severe COVID-19 [3], and patients weighting up to 150kg are common in ICUs. Obesity makes it even more difficult and physically demanding for caregivers to manipulate the patient's body when he/she is curarized.

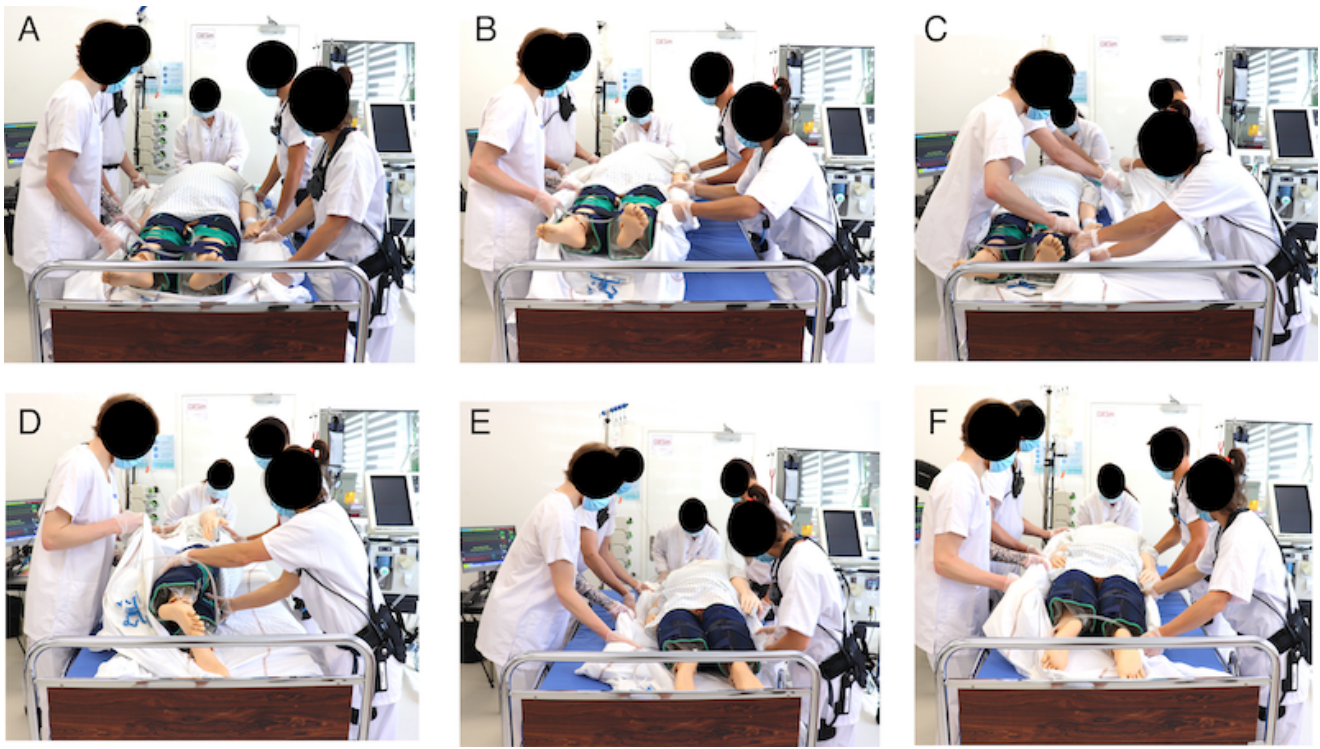

**Fig. 1.** The PP maneuver executed on a patient simulator (a manikin) at the Simulation Center of the University Hospital of Nancy (France). Although small differences in the practice of PP can happen from one hospital to another, the maneuver is substantially standardized (video tutorial: https://www.youtube.com/watch?v=X-qyeN3e8iU). *A*: A doctor positioned behind the head of the patient secure the head to avoid extubation and coordinate the whole procedure. Four teammates are distributed on both sides of the patient to reposition and turn the patient. *B:* The 4 teammates lift the patient and position him/her on one side of the bed. *C:* The 2 teammates on the right place a clean bed linen. *D:* The 2 teammates on the left pivot the patient temporarily on the side using the old linen. *E*: The patient is rotated toward vent until in prone position, lying on his/her abdomen. *F*: The 4 teammates lift the patient to position him/her at the center of the bed and add pillow underneath him/her.

Musculoskeletal injuries and back pain generated by the repetition of strenuous tasks are well-known in industrial scenarios [2], and many back-support systems have been proposed to alleviate the problem. Given the similarity of the postures, robotics assistance used in industry might also be useful to assist caregivers. In the specific case of PP maneuver, collaborative robots or mobile manipulators are not suitable, mainly due to limited robot payload and lack of available space in the ICU. Motorized beds that can help with manual repositioning of patients (e.g., *Hospidex toto* with inflatable air cells and *Vendlet V5S* with motorized bars) do exist, but they cannot fully replace the work of the caregivers and require substantial time and financial resources to be put in place. Conversely, occupational exoskeletons [4] appear as less invasive for the current practice, easy to set up, cheap, and compatible with the caregivers' work in the ICU.
Exoskeletons are wearable devices generally aimed at supporting physical tasks by generating appropriate force on one or multiple human joints. Recently, there has been increasing interest in employing exoskeletons for workplace ergonomics, to reduce physical workload [4] and risks of developing musculoskeletal disorders [5]. Exoskeletons can be active (motorized) or passive, in which case mechanical elements such as springs store and restore energy, transferring the load from one body part to another. Previous studies generally agree on the efficiency of passive back-support exoskeletons (designed to assist with the back and hip muscles) to reduce lumbar muscular activity and perceived exertion/discomfort, particularly during operations involving trunk flexion/extension in the sagittal plane [5].
However, the magnitudes of the reduction observed in back muscle activity when using those systems differ substantially from one study to another [6,7,8,9], likely because of considerable disparities in protocols (e.g. populations, tasks, postures, exoskeleton designs). This results in confusion regarding the magnitude of assistance to be expected when using a back-support exoskeleton. Besides, while occupational exoskeletons are deployed in the industrial sector [4], there are very few reports of their use in the healthcare system and no reported use in an ICU. Two active back-support exoskeletons (respectively the HAL for Care Support [10] and a custom prototype [11]) were tested in a simulated sit-to-stand patient transfer task, but the movements and postures involved differ strongly from those observed during PP maneuvers. The use of exoskeletons in the healthcare domain was also investigated in [12] to assist surgeons, but only focused on shoulder assistance.

Given the task-specific efficiency and functionality of exoskeletons [6], a multidisciplinary team of medical doctors, robotics researchers and ergonomists conducted a pilot study to evaluate the potential and feasibility of using occupational back-support exoskeletons to help caregivers in ICU during PP maneuvers. In this paper, we report on our pilot study that enabled to use the Laevo passive exoskeleton (Laevo, Ryswick, the Netherlands) in the ICU of the University Hospital of Nancy. The study was conducted in two steps. Initially, in a simulated environment, to probe into the helpfulness of an exoskeleton during PP. Then, in the real ICU to demonstrate the feasibility of using an exoskeleton during a COVID-19 PP shift. In this paper, we present an extended version of the preliminary work we presented in [13] accounting for feedback of use of the device from April 2020 to January 2021. Specifically, we report the experimental procedure and all the materials that we designed ad-hoc for this study, to make it fully reproducible by other robotics and medical teams. We report the entire questionnaires and further detail about the materials and methods in the Supplementary Material at the end of the article.

## 2    Exploratory Study to Select the Most Suitable Exoskeleton at the Hospital Simulation Center

Preliminary visual video analysis of PP maneuvers revealed that the medical staff can assume postures with forward trunk bending up to 45 degrees with raised arms straight forward, exert traction to the trunk bending up to 20-30 degrees, and hold prolonged static postures with the trunk bent forward up to 60 degrees. To confirm the visual observation, we recorded the whole-body kinematics of one physician (M, 35 years old, 175cm) performing the PP maneuver, using the Xsens MVN inertial motion capture system (Xsens, Enschede, the Netherlands, capture rate: 240Hz). Postural analysis with the AnyBody biomechanical simulation software (AnyBody Technology, Aalborg, Denmark) revealed that when operating at the side of the patient, the physician spends approximately 40% of the maneuver time with the torso bent more than 20 degrees forward; when operating behind the head of the patient to secure the head and avoid extubation, the physician maintains a static posture with important flexion of the trunk for several minutes. The precise angle of flexion, in this case, depends on the height of the patient's bed, his/her location relative to the bed, and the doctor's height. Even when not associated with load manipulation, such postures cause mechanical load in the lower back [2]. For such postures, exoskeletons for lumbar support can help, but to be used in ICU they have to match many usability constraints, such as being lightweight and unburdensome. Based on this analysis, and with the urgency of the situation in mind, we identified four commercial exoskeletons that were already available in our teams and that could meet the requirements: Corfor (Corfor, France), Laevo v1 (Laevo, Netherlands), BackX (SuitX, USA), and CrayX (German Bionics, Germany). Corfor is a passive soft exoskeleton (also known as *exosuit*), Laevo and BackX are passive rigid exoskeletons based on springs, while CrayX is an active exoskeleton, employing electrical actuators.

**Methods -** Five experienced PPT volunteers performed 11 PP maneuvers, consisting of Prone to Supine (PS) and Supine to Prone (SP) gestures, with a 100kg manikin. All participants filled a questionnaire to retrieve human factors (Questionnaire A, Supplementary Materials). Two of the participants (M, 30 and 35 years old) alternatively tried all 4 exoskeletons at the manikin's side. After testing each exoskeleton, these 2 subjects answered an evaluation and acceptance questionnaire (Questionnaire B, Supplementary Material) adapted from [14], consisting of 19 items on a 5-point Likert scale with a neutral option, to evaluate the perceived effort, safety, comfort, efficacy, installation, intention to use. They also reported on their perceived efforts using an effort evaluation questionnaire (Questionnaire C, Supplementary Material) and on their experience in a semi-directed interview conducted by one experimenter. To analyze the kinematics and dynamics of the maneuver, we replayed the participant's whole-body motion recorded with the Xsens motion capture system with a Digital Human Model (DHM) using the Dart physics engine. The DHM has 43 degrees of freedom and it is scaled based on the participant's height and mass using average anthropometric coefficients. We used a hierarchical velocity quadratic programming (QP) controller based on the OpenSoT library [15] to retarget the participant's upper-body motion directly in the Cartesian space (the lower-body was mainly fixed during the PP maneuvers).

In the absence of physiological measures, we used the DHM computed joint torques as a surrogate measure of the efforts performed by the participants during the experiment to investigate the effect of the exoskeleton. We used the DHM L5/S1 flexion/extension joint torques estimated with the dynamic simulation to compare the lumbar effort exerted by the participant with and without the exoskeleton. When no exoskeleton is worn, the joint torque exerted by the human is directly retrieved from the simulation. However, when the participant is equipped with the exoskeleton, the net torque exerted at the L5/S1 joint to counter the dynamics and gravity effects on the upper-body is a sum of the human-generated torque and of the exoskeleton assistive torque: $\tau_{L5S1} = \tau_{human} + \tau_{exo}$. The net torque is retrieved from the DHM simulation, but in order to estimate the human torque, the assistive torque $\tau_{exo}$ provided by each exoskeleton is needed. To compute this torque, one needs the details about the mechatronics design of the platform. Since the Laevo was unanimously perceived as the most suitable exoskeleton to help in PP maneuvers, we computed $\tau_{exo}$ for the Laevo only, using the Laevo empirical calibration curve published by Koopman et al. in [9] and the maximum torque (40 Nm) reported in the Laevo user manual. The dynamics analysis was performed with a simplified simulation, which has limitations. The effects of the Laevo weight (~2.5kg) and of external loads were ignored in the torque estimation. That is, we ignored the efforts associated with manipulating the manikin, and only focused on the postural efforts. This simplification matters only at the lateral position, since the external load at the head position was only associated with maintaining the endotracheal tube, and was not significant.

**Results -** In comparison with the non-assisted situation, both participants perceived a reduction of physical effort for all exoskeletons except the CORFOR. Table 1 reports the questionnaire's scores (*mean±stdev*) for each construct. All the exoskeletons scored positively in terms of perceived safety and comfort, and the participants did not notice a change in their performance, positive or negative, while using the exoskeleton. Laevo was the easiest to install and had the highest and only positive score in the intention to use. Both participants reported that CrayX was too cumbersome to wear in an ICU, while the mechanical design of BackX was unpleasantly hindering several arm movements of the PP maneuver. Both were restricting the range of motion for some gestures. CORFOR was not helpful. Conversely, they were satisfied with Laevo in terms of perceived assistance during bent postures, easiness of equipment, and freedom of movements. Importantly, they mentioned that Laevo was not modifying their movements during the PP maneuver, which was confirmed by the analysis of the kinematic data [13].

**Table 1**: Questionnaires scores for the 4 exoskeletons tested in the Simulation Center.

|  | Corfor | Laevo | BackX | CrayX |
|---|---|---|---|---|
| Reduction of physical effort | 3.0±0.0 | 4.0±0.0 | 4.0±0.0 | 4.0±0.0 |
| Perceived safety and comfort | 4.37±0.7 | 4.5±0.5 | 4.0±1.1 | 3.8±1.0 |
| Easiness to install | 3.5±2.1 | 4.5±0.7 | 1.5±0.7 | 1.5±0.7 |
| Intention to use | 3.0±0.0 | 4.5±0.7 | 2.5±0.7 | 3.0±0.0 |

**Kinematic analysis -** We first verified whether the motion of the L5/S1 joint in the sagittal plane (low-back flexion/extension motion) was affected by the use of any of the 4 exoskeletons. We manually segmented each trial into 2 segments: Prone to Supine (PS) and Supine to Prone (SP) positioning. We compared the low-back flexion angle for the PS and SP, for all 4 exoskeletons and without exoskeleton: the joint angle profiles are overall similar for all conditions; their minimum and maximum values are similar (typical profiles in Supplementary Material). The median value of the back-flexion angle of the participant across one trial did not vary significantly from one condition to another (Fig. 3). Minor variations from one condition to another can be explained by small differences in the manikin's position on the bed and the overall maneuver performed by the team. These results suggest that the range of motion of the L5/S1 flexion/extension joint during the PP maneuver was not affected by the use of any of the exoskeletons.

**Dynamic analysis -** We estimated the human L5/S1 flexion torque during the PP maneuver, both with and without the Laevo (Fig. 4). When using Laevo for the SP and PS maneuvers, the low back torque medians were reduced by 11.3% and 13.0% respectively. Those results, though limited to one participant, suggest that wearing the Laevo may reduce the human low-back torque during PP maneuvers, which agrees with the subjective report of the participants. This result also agrees with the 15% reduction of the L5/S1 moment observed in [9] for a similar static forward bending task.

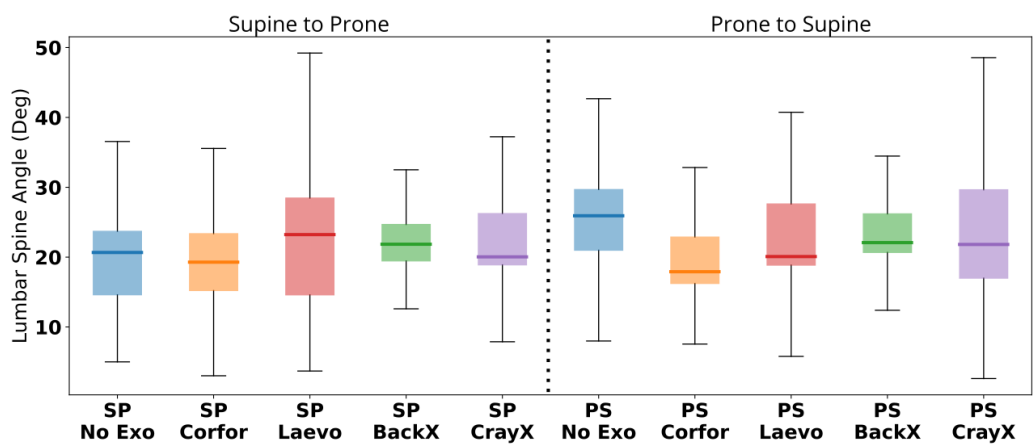

**Fig. 3.** Lumbar spine flexion angle of participant 1 performing the PP maneuvers. Boxplots represent the distribution of angle across time, both for the prone to supine (PS), and supine to prone (SP) maneuvers.

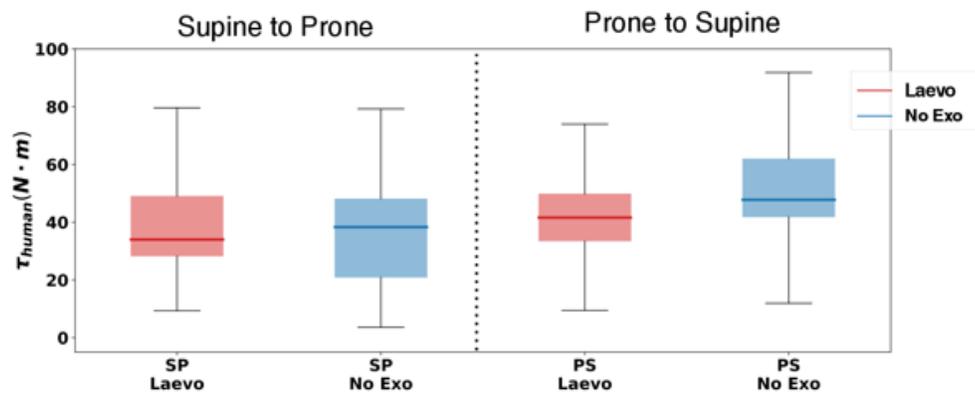

**Fig. 4.** Lumbar flexion torque of participant 1 performing the PP maneuvers. Boxplots compare the torque at L5/S1 with and without the Laevo, estimated with the DHM simulation.

## 3  Quantification of the LAEVO's assistance at the hospital simulation center

To complement the preliminary results described in the previous section, we performed a subsequent experiment to probe into the physiological effects of the Laevo during PP maneuvers. For sanitary reasons, this experiment was also conducted at the Hospital Simulation Center.

**Methods -** The same 2 volunteers were equipped with an ECG sensor (Delsys Trigno ECG Biofeedback, 2 channels, bandwidth: 30Hz, sampling rate 4370 sa/sec with onboard Butterworth bandpass filter 40/80 dB/Dec) and 12 surface EMG sensors (Delsys Trigno, EMG sampling rate 4370 sa/sec). Surface electrodes were positioned on the skin according to Seniam protocol recommendations after abrasion and cleaning with alcohol. The muscle activity was recorded bilaterally in erector spinae longissimus (ESL), erector spinae iliocostalis (ESI), trapezius ascendens (TA), biceps femoris long head (BF); and laterally on the right side for rectus abdominis at T10 level (RA), rectus femoris (RF), gluteus maximus (GM) and tibialis anterior (TAL). The participants performed the PP maneuver without and with the Laevo exoskeleton, positioned successively at the head and at the side of the manikin. The physiological signals were also recorded during a control condition, where the subject was standing still and silent. For all the conditions, we recorded a single repetition lasting 4 minutes.

**Results -** *Heart rate:* The instantaneous heart rate was computed offline by the Delsys EMGWorks software with its proprietary Template Matching method. Fig. 6 compares the instantaneous heart rate across conditions for the first participant. Table 2 summarizes the median heart rate value for both participants. Overall, we did not observe any significant change in the heart rate when using the exoskeleton. For participant 1, we observed a slight decrease in the median heart rate value in both positions (head: -3.66%; side: -6.07%), whereas for participant 2 the median heart rate either increased (head: +4.85%) or decreased (side: -6.61%). It must be noted that both participants are confirmed athletes, which explains their low basal heart rates. The weak differences in heart rate do not indicate a particular physiological gain when using the Laevo for such short tasks (4 minutes). For longer tasks and/or more repetitions of PP maneuvers results may differ, and a more thorough evaluation is needed to confirm.

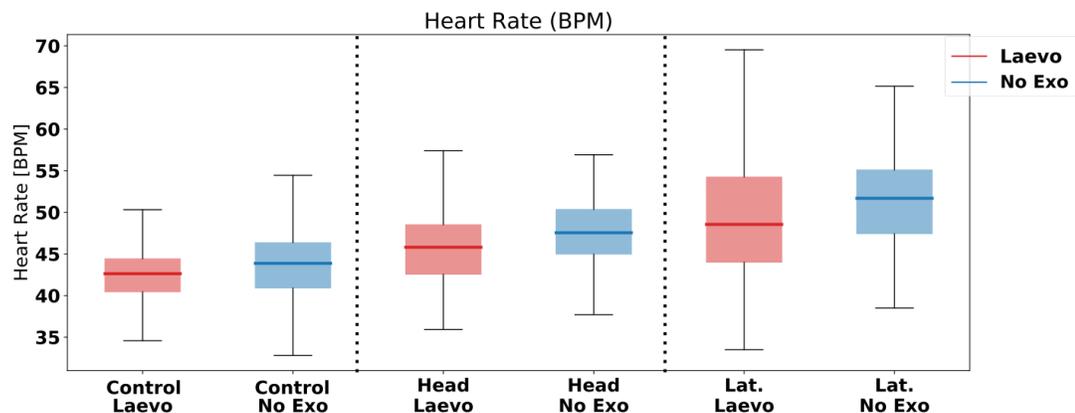

**Fig. 6.** Boxplots representing the distribution across time of the heart rate of participant 1 with and without the Laevo, measured during a control resting condition and while executing the PP maneuver at the head and at the side of the simulated patient.

**Table 2:** Median across time of heart rate value (in bpm) for the 2 participants (P1 and P2) performing a PP with and without the Laevo exoskeleton.

|     | Head No Exo | Head Laevo | Side No Exo | Side Laevo |
| --- | --- | --- | --- | --- |
| P 1 | 47.55 | 45.80 | 51.69 | 48.55 |
| P2  | 64.88 | 68.03 | 71.95 | 67.19 |

*EMG:* The raw EMG signal was rectified and filtered through an RMS filter with a 100ms window, then by a 4th order Butterworth low pass filter with a 10Hz cut-off frequency. Table 3 displays the change in the RMS value of the EMG signal when using the Laevo, compared to the baseline with no exoskeleton. We observed a general decrease of the activity of the back-extensor muscles (except ESI which was not affected and the head condition for P2) and the hip extensor muscles, when using the Laevo, probably depending on the task, the position in the team and the individual characteristics. Conversely, the activity of the antagonist trunk and leg muscles and of the lower-leg muscles remained similar when using the Laevo. Although our results are strongly limited by the small number of participants and the absence of repetitions of the task, they suggest a promising physiological gain brought by the Laevo. This gain is coherent with the subjective evaluation retrieved in the questionnaires, and endorsed the use in the ICU. Furthermore, those results are in line with the changes in muscle activity observed in other studies where the Laevo was tested in similar static forward bending tasks [9,16].

**Table 3**: Changes in RMS value of the EMG signal when using the Laevo, compared to the baseline without exoskeleton, for both participants (P1 and P2) and both positions (in %).

|         | TA     | ESL    | ESI   | GM     | RF    | BF     | RA    | TAL    |
| ---     | ---    | ---    | ---   | ---    | ---   | ---    | ---   | ---    |
| P1 Head | -35.24 | -11.94 | -0.70 | -30.10 | +0.35 | -16.23 | -1.20 | -1.83  |
| P2 Head | +2.20  | +5.15  | +0.42 | +4.48  | NA    | -20.38 | -0.12 | +5.60  |
| P1 Side | -14.75 | -8.81  | -1.04 | -11.51 | -5.80 | -52.69 | +0.32 | -9.73  |
| P2 Side | -4.81  | -11.42 | -1.84 | +7.71  | NA    | -33.87 | -0.48 | -19.84 |

NA=not available because of a technical problem with the sensor reading.

## 4 Implementation in Real-life Conditions in the ICU of the University Hospital of Nancy

Building upon these promising pilot results, and given the urgency associated with the COVID-19 crisis, we proceeded to test in realistic conditions to demonstrate the feasibility of using Laevo in a COVID-19 ICU situation.

**Methods -** The same 2 participants were each equipped with a Laevo complying with the drastic hygiene rules of the ICU during the outbreak. Each participant wore the Laevo over a single-use scrub suit and underneath the personal protective equipment (long sleeve gown) covering the entire body to reduce the risk of contamination. That way the exoskeleton was not in contact with the patient nor with the contaminated environment. During a typical 3-hour shift, the participants performed 10 PP maneuvers on 10 ICU patients; each participant was positioned 3 times at the head and 7 times at the patient's side, respectively. After each maneuver, the participants reported their perceived effort using a Borg-CR10 scale (questionnaire D part 1, Supplementary Material). Cardiac activity was monitored with a Holter-ECG during the whole shift in the ICU. At the end of the shift, they reported on the use of the exoskeleton (questionnaire D part 2, Supplementary Material), then filled the extended version (22-items) of the exoskeleton evaluation questionnaire (questionnaire B, Supplementary Material), and reported on their experience in an interview. Concomitantly, the PPT colleagues filled a questionnaire to report on their experience working alongside people equipped with exoskeletons (questionnaire E, Supplementary Material).

**Results -** Both physicians perceived a physical relief in the low back during bent postures, particularly when working at the patient's head, and indicated an intention to adopt such technology after the pilot study. ECG data analysis was inconclusive due to the multiple biases in this real-life condition (e.g., elevated stress of the participants due to the COVID-19 context) and the frequency of multiple gestures performed during the PP maneuver [17]. Overall, Laevo was positively evaluated in terms of physical relief (questionnaire score: 4.2±0.4): both participants reported that the perceived general fatigue at the end of the shift was reduced when using Laevo. Importantly they both said they would use Laevo again without hesitation for future shifts where they would be positioned at the patient's head (effort score on Borg-CR10 scale: head: 1.8±0.4; side: 3.2±0.7). The participants found the Laevo comfortable (questionnaire score: 4.5±0.5), except when walking, which is a well-known issue of Laevo v1 [18] that was improved in recent versions. Laevo did not prevent or constrain their usual gestures and activity in the ICU (Fig. 7). Accordingly, their teammates did not notice any particular changes in the practice, and no physical or psychological side effects were observed. Those results are important for a future adoption of Laevo in the current practice, because the positive attitude of the co-workers is fundamental for the acceptance of new technology at work [19].

## 5   Use of the Exoskeleton in the ICU of the University Hospital of Nancy in the second wave of pandemic

Two additional Laevo v2.5 were provided to the medical staff, for a total of 4 exoskeletons since October 2020 (two v1 and two v2.5). At the present day (January 26, 2021) the 4 exoskeletons have been tested by more than 60 healthcare workers (medical doctors and nurses). Preliminary feedback from the medical staff confirms that the exoskeletons are useful to reduce the physical effort and back pain. Laevo v2.5 is preferred because of the walking feature (a lever enables to unblock the exoskeleton and walk normally). The adoption of the devices has been variable: some physicians are regular users (i.e., they ask to use the exoskeleton when they have to perform one or more PP), others are neutral or even sceptic.

The characteristics of the second wave of COVID-19, interestingly, have determined some differences in the use of the exoskeletons. The number of daily PP maneuvers performed in the ICU has been highly variable but generally lower than the in the first wave, sometimes even zero: so, the exoskeletons have not been used on a regular basis. The lower number of PP and the staff rotations made it possible for physicians to significantly reduce the number of PP per day: so, the exoskeleton was really not used as a personal equipment but shared among a pool of users and for few PP. While this could seem a positive point (it can help a larger group of users), it introduced a fundamental problem which is the repetitive act to calibrate and fit the exoskeleton to each person's body at each use. This caused a major obstacle to the regular use: several users reported that they considered it "difficult to adjust" and that the fitting took too much time. When physicians only have 1 PP to perform, for example, the benefit brought by the exoskeleton is not worth the time and frustration they experience when they have to wear it and adjust it. Interestingly, this problem is even bigger for Laevo v2.5, since the physicians must change the metallic bars corresponding to their size, an operation that is considered also difficult.

Overall, while it is clear that the exoskeleton is helping the physicians, its use as a shared tool (and not as a personal equipment) is preventing its adoption.

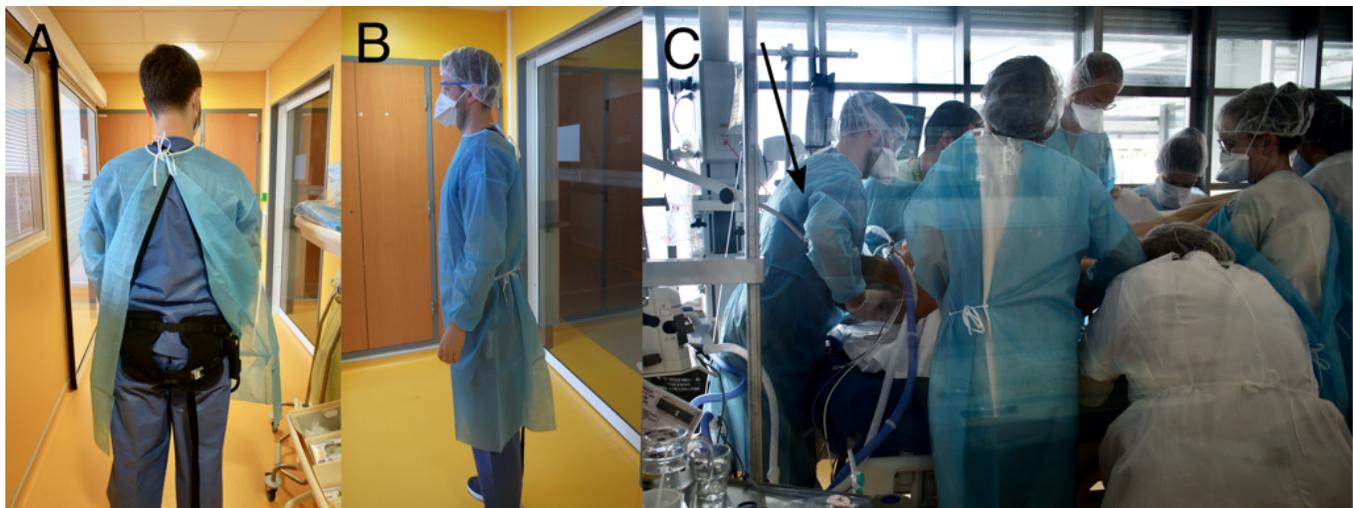

**Fig. 7.** Laevo used in the ICU. A and B: the exoskeleton is worn on clothes, hidden below the medical gown. C: the exoskeleton assists the physician positioned behind the head of the patient during a PP maneuver.

## 6   Conclusion

Our pilot study demonstrated that back-support exoskeletons are readily feasible to deploy in an ICU to physically assist the medical staff engaged in PP maneuvers, even in the dreadful sanitary context of the COVID-19 pandemic.
Despite the many limitations of our exploratory study, carried out rapidly due to the urgency of the situation and with limited availability of physicians heavily busy in the hospital, we were able to demonstrate the feasibility and potential usefulness of using a back-support exoskeleton to assist during PP maneuvers in the ICU in less than two weeks. Though only 2 physicians used the passive exoskeleton Laevo during the PP maneuvers in the ICU in April 2020, they both perceived a physical relief in the low back during bent postures, particularly when working at the patient's head, and indicated an intention to adopt such technology after the pilot study. EMG and ECG analysis in the lab study confirm the help of the device. Four Laevo exoskeletons have been used in the ICU of the Hospital of Nancy since October 2020: tested by more than 60 people, they have been used regularly only by a small group of physicians. Their adoption is currently mainly limited by the use as a shared tool, while they should be a personal equipment. This is not surprising: it is known that the performance and physical assistance provided by an exoskeleton are not the only criteria for its adoption, and other factors, individual or organizational, are also critical [14]. Nevertheless, the positive feedback from the regular users of the exoskeleton is very encouraging: we are currently evaluating the use of the exoskeletons in the ICU as a personal equipment with a small group of volunteers.

The future adoption of this technology in the current practice will require more thorough studies quantifying the physiological benefits for a larger participant group size, with more repetitions of the maneuver to match the real use of the exoskeleton over an entire work shift, and evaluating the effects in a longitudinal study.

By reporting the experimental protocol and all the materials, we encourage other teams to replicate our study, help the medical staff of other hospitals, and to share feedback on experience among interested worldwide intensive care units.

# Supplementary Material

I. QUESTIONNAIRES

In this supplementary material, we report the entire set of questionnaires that we used during our study, to facilitate reproducibility and replication of our results by other teams.

## A. *Human factors related to prone positioning*

This questionnaire must be filled by **any participants** to the prone positioning (PP) study before doing any PP maneuver in simulation or in the ICU. It is necessary to retrieve relevant information such as experience with the PP in the hospital before and during the COVID-19 pandemic, history of back pain that is medically relevant for the study, attitude towards the exoskeleton technology. A semi-directed interview to gain a deeper knowledge of the expectations of the participants is complementary: the experimenter can ask, for example, if the participants have already seen or used an exoskeleton, a robot or a prosthetic device, and get insights into their past experience.

| Questionnaire on human factors | |
|---|---|
| Instructions: This questionnaire is to be completed once, before the start of any experiment. | |
| Question | Answer |
| Participant ID | |
| Gender (M/F) | |
| Age | |
| Occupation (nurse, doctor…) | |
| Number of months/years of hospital experience | |
| Number of days/months/years of experience performing the PP maneuver | |
| In the past, have you had back problems (recurring pain that requires medical attention or sick leave)? | |
| Do you currently have back problems? | |
| Have you ever used a back-support system? If yes, which one(s) | |
| Have you ever interacted with robotic systems or physical assistance devices such as exoskeletons? If yes, which one(s) | |
| What is your overall attitude towards physical assistive devices such as exoskeletons? | ❏ Very negative<br>❏ Rather negative<br>❏ No opinion<br>❏ Rather positive<br>❏ Very positive |
| How many times have you performed this maneuver in the field since the beginning of your activity as a caregiver? | ❏ 1-10<br>❏ 10-50<br>❏ 50-100<br>❏ 100+ |
| Do you consider yourself an expert in this maneuver? | ❏ Expert<br>❏ Medium<br>❏ Beginner |
| **Before COVID-19** | |
| On a scale of 1 (not stressful at all) to 10 (very stressful), if you performed the PP maneuver in the past, before the COVID-19 outbreak, how stressful was it …? | |
| … physically | ❏_______ |
| … cognitively | ❏_______ |
| How often did you execute the PP maneuver before the crisis situation COVID-19? | |
| … times per day | ❏_______ |
| … times per month | ❏_______ |
| **During COVID-19** | |
| On a scale of 1 (not stressful at all) to 10 (very stressful), if you performed the PP maneuver during the COVID-19 outbreak, how stressful was it …? | |
| … physically | ❏_______ |
| … cognitively | ❏_______ |
| How often do you execute the PP maneuver every day during the COVID-19 outbreak? | |
| … times per day | ❏_______ |

| | | |
|---|---|---|
| | … times per month | ❏ _______ |

## B. Acceptance evaluation of an exoskeleton for prone positioning

This questionnaire is an extract from a larger questionnaire[1] that was developed by INRS to investigate the acceptance of exoskeletons introduced in an industrial context. Some questions have been adapted to our specific use case.

This questionnaire must be filled by the **physicians** equipped with the exoskeleton to receive physical assistance during the prone positioning maneuver. It is used to evaluate the exoskeletons in the simulated study and in the real-life conditions. Questions marked with an asterisk are to be filled only after using the exoskeleton in real-life condition, i.e., in the ICU. Reverse questions are marked with R.

| Questionnaire for exoskeleton evaluation | | | |
|---|---|---|---|
| Question | | | Answer |
| Participant ID | | | |
| Exoskeleton type | | | ❏ Laevo<br>❏ Corfor<br>❏ CrayX<br>❏ BackX |
| Instructions: In the questionnaire you will find a series of statements about your experience with the exoskeleton. For each statement that follows, please give your opinion by checking the corresponding box on a five-point Likert scale.<br>Scale A:<br>  1. Strongly disagree<br>  2. Disagree<br>  3. Neither Agree nor Disagree<br>  4. Agree<br>  5. Strongly agree<br>Scale B:<br>  1. Much lower<br>  2. Lower<br>  3. Identical<br>  4. Higher<br>  5. Much higher | | | |
| N. | Reverse & ICU | Question | Scale |
| | | Exoskeleton setup and calibration | |
| 1 | | I find the exoskeleton is easy to set up | A |
| | | The use of the exoskeleton | |
| 2 | | Overall, I find the exoskeleton easy to use | A |
| 3 | | I find that I can easily perform my movements with the exoskeleton | A |
| 4 | | I find that I can easily move and walk with the exoskeleton | A |
| 5 | | I find that I control my gestures as I wish with the exoskeleton | A |
| 6 | R | I find that the exoskeleton prevents me from working the way I want | A |
| 7 | | I find that I easily got used to working with the exoskeleton | A |
| 8 | R | I find that using the exoskeleton requires an extra effort of concentration | A |

---

[1] Wioland L., L. Debay, J.-J. Atain-Kouadio (2019) Processus d'acceptabilité et d'acceptation des exosquelettes: évaluation par questionnaires. Références en santé au travail, TF 274, n. 160, pp. 49 - 76. Available at: http://www.inrs.fr/dms/inrs/CataloguePapier/DMT/TI-TF-274/tf274.pdf

| | | | |
|---|---|---|---|
| | | My performance with the exoskeleton | |
| 9 | | I find that the speed of my work with the exoskeleton is … | B |
| 10 | * | I find that the quality of my work with the exoskeleton is … | B |
| 11 | | I find that my effectiveness with the exoskeleton is … | B |
| 12 | | I find that the productivity of the team with the exoskeleton is … | B |
| | | My health and safety | |
| 13 | R | Overall, I find that my physical efforts with the exoskeleton are … | B |
| 14 | R | Overall, I find that with the exoskeleton, my fatigue is ... | B |
| 15 | | I feel safe working with the exoskeleton. | A |
| | | My feeling with the exoskeleton | |
| 16 | R | I feel nervous when I use the exoskeleton. | A |
| 17 | R | I feel worried when I use the exoskeleton. | A |
| 18 | | I feel confident when I use the exoskeleton. | A |
| 19 | R | I find I annoy my colleagues when I use the exoskeleton. | A |
| | | Future use | |
| 20 | | If I have a choice, I am thinking of using or continuing to use the exoskeleton in the next months | A |
| 21 | * | I find that over the course of the day I have adapted to the exoskeleton | A |
| 22 | * | I find that using the exoskeleton during the day has been beneficial | A |

*C. Evaluating the overall effort of a prone positioning maneuver with/without an exoskeleton*

This questionnaire must be filled after realizing a prone positioning (PP) maneuver with or without an exoskeleton. It provides a subjective evaluation of the amount of physical effort and discomfort perceived while executing the PP. It provides a surrogate measure of standardized quantitative measures of efforts, such as surface EMGs placed over muscles of interests, whenever obtaining such measures is not possible. For example, using surface EMGs in the ICU was not possible for sanitary reasons.

| Questionnaire for evaluation of the overall effort | |
|---|---|
| Question | Answer |
| Participant ID | |
| Exoskeleton type | ❏ Laevo<br>❏ Corfor<br>❏ CrayX<br>❏ BackX<br>❏ none |
| Instructions: This questionnaire is to be completed after a PP maneuver with or without an exoskeleton. Put a circle on the relevant areas in the double-sided images, as well as a number, as per instructions. | |
| Image A: 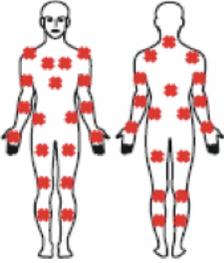 | |
| Image B: 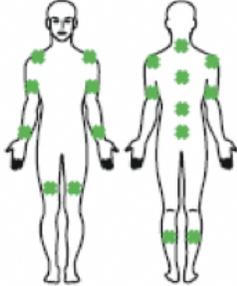 | |

| Question | Image |
|---|---|
| Put a circle on this image on the areas where you felt physical effort during PP maneuvers [with the exoskeleton]. | A |
| Put a circle on this image on the areas where you felt physical effort becoming annoying during the PP maneuver [with the use of the exoskeleton], and a number next to each circle (1-10) to indicate the severity (1= no effort, 10= a lot of effort). | A |
| Only if you used the exoskeleton | |
| Could you indicate on this image the areas where you feel a new redistribution of physical effort, which you do not feel without the exoskeleton? Put a circle on the areas where you feel new efforts, and a number next to each circle (1-10) to indicate how much (1= no effort, 10= a lot of effort). | B |
| Could you indicate on this image the areas where you feel discomfort caused by the exoskeleton? Put a circle over the areas of discomfort, and a number next to each circle (1-10) to indicate how much (1= no discomfort, 10= a lot of discomfort). | A |

*D. Questionnaire for physicians using the exoskeleton in the ICU*

This questionnaire must be filled by the **physicians** equipped with the exoskeleton and using them in the ICU for PP maneuvers. The first part must be filled at the end of each PP maneuver (i.e., if the physician performs 20 maneuvers, he/she will fill the sheet 20 times, one after each PP). The second part must be filled at the end of the work-shift, for example in the changing room of the medical staff to avoid contamination. It must be noted that since the first batch of questionnaires are filled in the ICU following each PP, the paper sheets are considered "potentially contaminated" and therefore should rest untouched in a safe place for 3-4 days before manually processing them by the experimenters, in order to reduce any risk of contamination by direct skin contact.

| |
|---|
| **Questionnaire to evaluate the exoskeleton in the ICU** |
| Instructions: This questionnaire is to be completed by volunteers equipped with exoskeletons throughout their working hours. The first part must be filled after each PP maneuver. The second part must be filled at the end of their working hours. |
| **1st PART : TO FILL AFTER EVERY PP** |

| Question | Answer |
|---|---|
| Participant ID | |
| n. PP of the day | |
| With exoskeleton? | Yes / No |
| Position? | Head / Side |
| Your perception during the PP maneuver ||
| Please note for each anatomical zone your perceived effort during your last PP maneuver according to Borg's Perceived Effort Rating Scale (Borg CR10).<br><br>0   nothing at all<br>0.5 extremely weak / very, very slight<br>1   very weak / very slight<br>2   weak / slight<br>3   moderate<br>4<br>5   strong / severe<br>6<br>7   very strong / very severe<br>8<br>9<br>10 extremely strong / very, very severe ||
| Neck | ❏ _______ |
| Lower back | ❏ _______ |
| Legs | ❏ _______ |
| Left shoulder / arm | ❏ _______ |
| Left forearm / hand | ❏ _______ |
| Right shoulder / arm | ❏ _______ |
| Right forearm / hand | ❏ _______ |
| **2nd PART: FILL WHEN YOUR WORK WITH THE EXOSKELETON IS FINISHED** ||
| Question | Answer |
| Participant ID | |
| Exoskeleton type | ❏ LAEVO |
| Between when you started to work and now, how many times have you practiced the PP maneuver today? | |
| Have you systematically used the exoskeleton to perform the PP maneuver? | Yes / No |
| If not, why did you remove it? | |
| In how many of the total PP maneuvers out of the total PP maneuvers did you use the exoskeleton? | _______ *(number)* out of _________ *(total)* |
| In total, how long did you keep it? | |
| Did you change any settings during use? If yes, specify when and why. | |
| Did you change any settings after removing the system (for example, after using the restroom)? If so, specify when and why. | |
| Did you unhook the thighs pads to walk? If yes, how many times? | |
| Did the exoskeleton prevent you from making one or more movements? If yes, can you list them. | |
| Did something unexpected happen? If so, can you describe it? | |
| Do you have any comments about your experience today as a physician equipped with an exoskeleton? Are there things you noticed while using and working with the exoskeleton? | *(free comment)* |

### E. *Questionnaire for colleagues in the ICU*

This questionnaire must be **filled by colleagues of the physicians equipped with the exoskeleton**. The first part is filled before the normal work-shift. The second part is filled at the end of the work-shift, when they have had the experience of working with colleagues wearing the exoskeleton. The questionnaire can be filled outside the ICU, for example in the medical staff changing rooms, thus avoiding any risk of contamination by paper.

| **Questionnaire for colleagues in the ICU** | |
|---|---|
| Instructions: you are going to work alongside people equipped with exoskeletons. We need to get some information about your perceptions. This is important for our research to evaluate the impact of introducing such a tool in a hospital setting during COVID-19. The questionnaire consists of two parts: the first, to be completed just before you start working with your colleagues equipped with exoskeletons in the ICU; the second, at the end of your working day. Please fill the two parts outside the ICU. | |
| **1st PART: TO BE FILLED BEFORE ANY WORK WITH PEOPLE USING EXOSKELETONS** | |
| Question | Answer |
| Participant ID | |
| Gender (M/F) | |
| Age | |
| Occupation (nurse, doctor...) | |
| Number of months/years of hospital experience | |
| Number of days/months/years of experience performing the PP maneuver | |
| In the past, have you had back problems (recurring pain that requires medical attention or sick leave)? | |
| Do you currently have back problems? | |
| Have you ever used a back-support system? If yes, which one(s) | |
| Have you ever interacted with robotic systems or physical assistance devices such as exoskeletons? If yes, which one(s) | |
| What is your overall attitude towards physical assistive devices such as exoskeletons? | ❏ Very negative<br>❏ Rather negative<br>❏ No opinion<br>❏ Rather positive<br>❏ Very positive |
| **2nd PART: TO BE FILLED AFTER YOU WORKED WITH PEOPLE USING EXOSKELETONS** | |
| You have worked with people wearing an exoskeleton. | |
| Did working next to a colleague with an exoskeleton make you nervous? | ❏ 1 = not at all<br>❏ ..<br>❏ 10 = very nervous<br><br>Answer: _______ |
| Have you been annoyed by working next to people with exoskeletons? | ❏ 1 = not at all |
| | ❏ .. <br> ❏ 10 = very annoyed <br><br> Answer: _________ |
| Compared to the "normal" situation (no exoskeleton), did you find the new situation more physically demanding? | ❏ 1 = much less <br> ❏ … <br> ❏ 5 = identical <br> ❏ … <br> ❏ 10 = much more <br><br> Answer: _________ |
| Compared to the "normal" situation (no exoskeleton), did you find the new situation more cognitively demanding (for example, you had to pay more attention...)? | ❏ 1 = much less <br> ❏ … <br> ❏ 5 = identical <br> ❏ … <br> ❏ 10 = much more <br> ❏ <br><br> Answer: _________ |
| If you had the choice, would you use an exoskeleton yourself in the next few months if the current sanitary situation continued? | ❏ Strongly disagree <br> ❏ Disagree <br> ❏ Neither agree nor disagree <br> ❏ Agree <br> ❏ Strongly agree |
| **Your perception about the PP maneuver during COVID-19** | |
| On a scale of 1 (not stressful at all) to 10 (very stressful), if you performed the PP maneuver during the COVID-19 outbreak, how much "stressful" was it …? | |
| … physically | ❏_______ |
| … cognitively (pay attention) | ❏_______ |
| How often do you practice the PP maneuver every day during the COVID-19 outbreak? | |
| … times per day | ❏_______ |
| … times per month | ❏_______ |
| How many times have you practiced the PP maneuver today? | |
| Do you have any comments on your experience today working alongside people equipped with exoskeletons? <br> Are there things you noticed about the use of and the work with the exoskeleton? | *(free comment)* |
| Please note for each anatomical zone your perceived effort during your last PP maneuver according to Borg's Perceived Effort Rating Scale (Borg CR10). <br><br> 0   nothing at all <br> 0.5 extremely weak / very, very slight <br> 1   very weak / very slight <br> 2   weak / slight <br> 3   moderate <br> 4 <br> 5   strong / severe <br> 6 <br> 7   very strong / very severe <br> 8 <br> 9 <br> 10 extremely strong / very, very severe | |
| Neck | ❏_______ |
| Lower back | ❏_______ |
| Legs | ❏_______ |
| Left shoulder / arm | ❏_______ |
| Left forearm / hand | ☐______ |
| Right shoulder / arm | ☐______ |
| Right forearm / hand | ☐______ |

II. EXOSKELETONS USED IN THE EXPLORATORY STUDY

Four commercial exoskeletons were used in the exploratory study at the Hospital Simulation Center: Corfor (Corfor, France), Laevo v1 (Laevo, Netherlands), BackX (SuitX, USA), and CrayX (German Bionics, Germany). It must be noted that in this study we are not claiming a comprehensive comparison of the different exoskeletons that could have been helpful. At the time of the study (which was during the 1st wave of the COVID-19 pandemic) it was not possible for us to rapidly purchase more exoskeletons to evaluate, and it is possible that other passive or active exoskeletons for low-back support which exist on the market could have been helpful for our task. Rather, we have an empirical proof of feasibility at least for one exoskeleton: it can be used in the ICU to assist in Prone Positioning, it is well perceived and possibly beneficial.

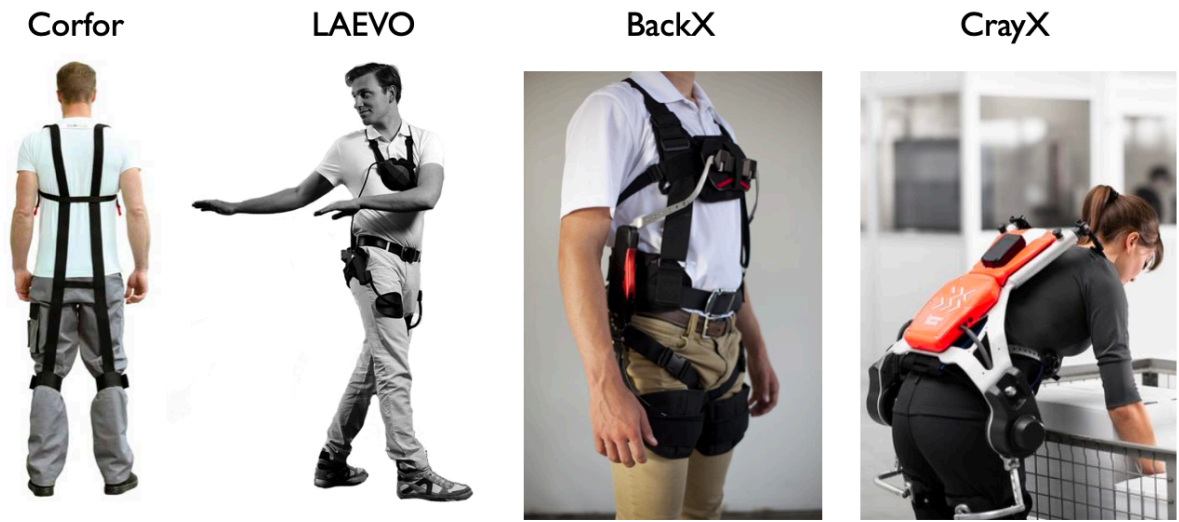

*Figure 1: the four commercial exoskeletons used in the exploratory study*

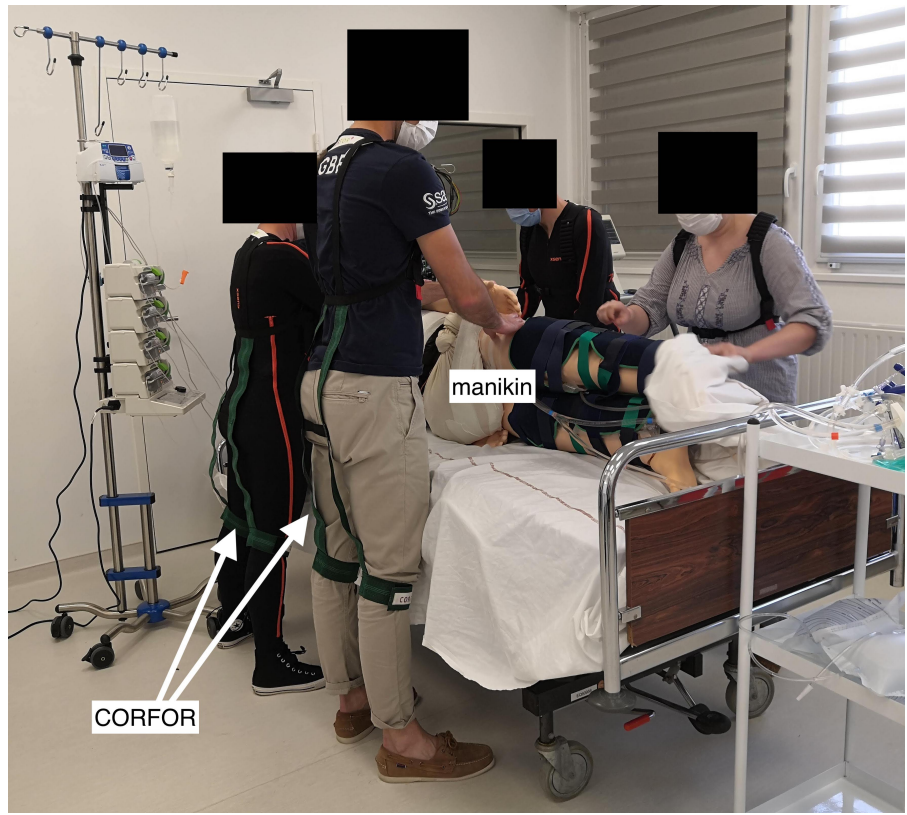

*Figure 2 - evaluation of the Corfor exosuit by the PP team. We assigned a Corfor system to each participant, choosing the size according to their height and the recommendations of the manufacturer. Two participants are also equipped with the Xsens MVN suit to record their motion. The device was reported as not helpful for the specific PP maneuver.*

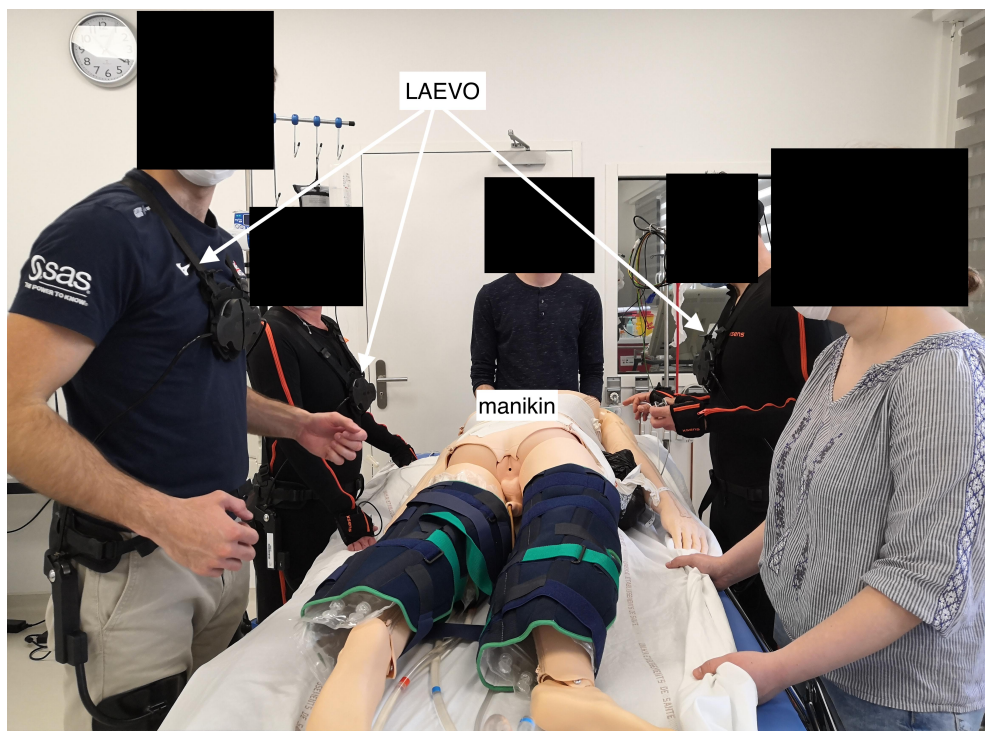

*Figure 3 - evaluation of the Laevo exoskeleton by the PP team. We only had three Laevo v1, one for each size (small, medium, large). Each exoskeleton was attributed to the participants according to their height and following the recommendations of the manufacturer. The Laevo exoskeleton was immediately perceived helpful and intuitive. One participant reported a slight discomfort on the sternum during back flexion and on the thighs during walking, noticing it would be better to unlock it to walk normally. This issue was solved in later versions of the Laevo exoskeleton, such as the v2.5 that was purchased and used in the 2$^{nd}$ wave of the pandemic in the ICU (since October 2020).*

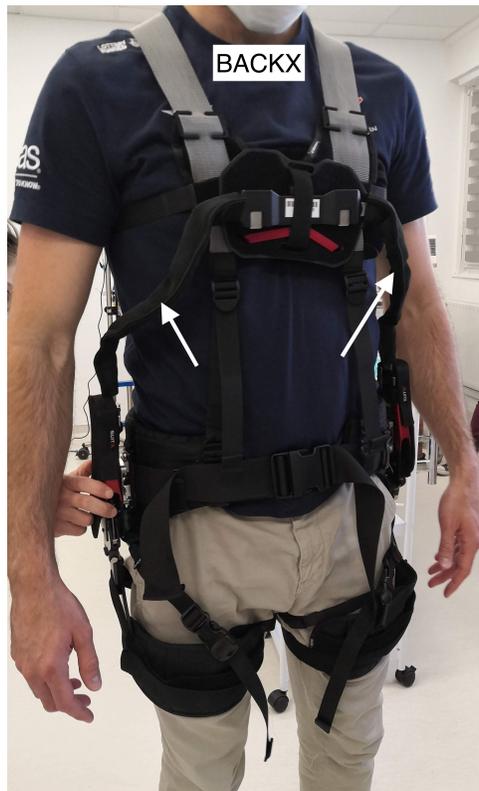

*Figure 4 - BackX worn by one participant. The arrows point to the metallic arcs that are constraining the arms during the PP gestures: the participants pointed out that this was one of the main reasons for not choosing to use this exoskeleton for this particular gesture. The BackX was perceived similar to the LAEVO in terms of assistance, but both participants reported that the metallic curved bars from the sternum to the hip were preventing several arm movements necessary to complete the PP maneuver, and as such they felt they could not execute the entire maneuver with this equipment in real conditions.*

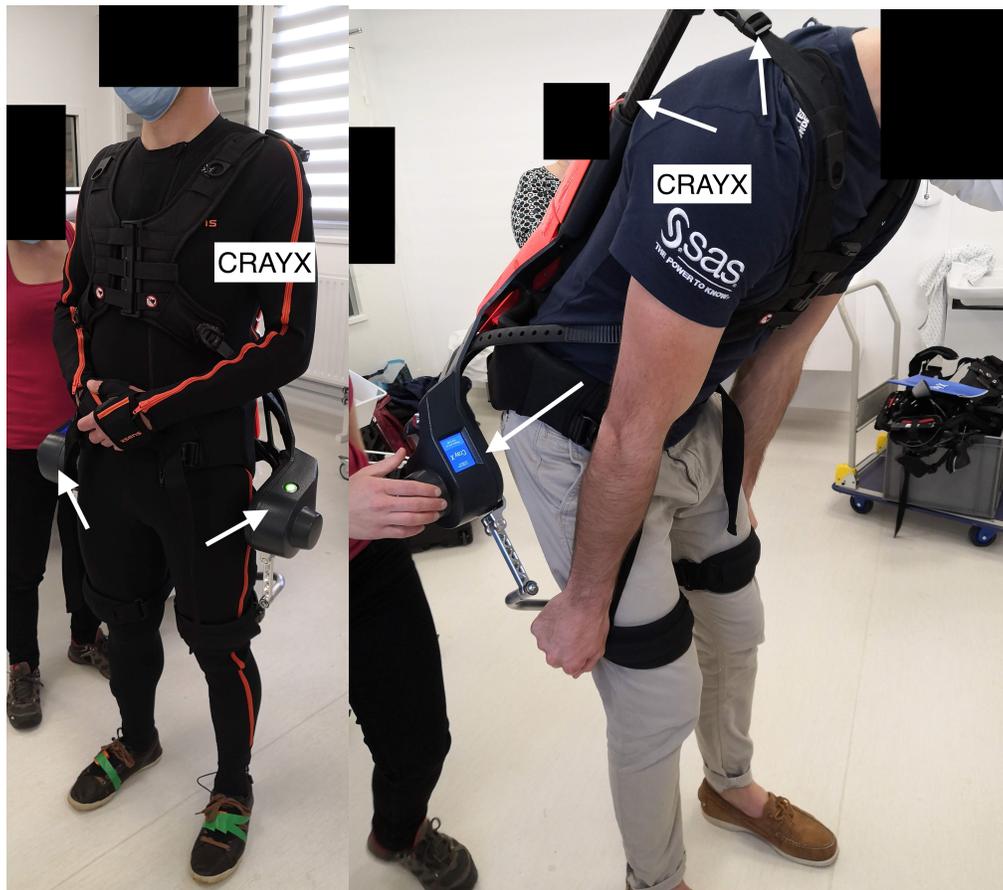

*Figure 5 - CrayX worn by two participants. The arrows highlight the external parts of the exoskeleton that add constraints to the workspace. The CrayX was considered too cumbersome to be used around patients, and very difficult to tune. Both participants perceived it required additional concentration effort during back flexion to enable the active support, which they couldn't improve even after changing the sensitivity parameter. Furthermore, one participant reported a critical discomfort due to a part of the CrayX applying a force on the dorsum.*

## III. Analysis of motion and lumbar effort with Digital Human Model

We replayed the participant's whole-body motion recorded with the Xsens motion capture system with a 43 DoFs Digital Human Model (DHM) using the Dart physics engine. The DHM consists of 19 rigid bodies linked together by 18 compound joints, for a total of 43 DoFs (11 for the back and neck, 9 for each arm including the sternoclavicular joint, and 7 for each leg), plus 6 DoFs for the free-floating base. Each DoF is a revolute joint controlled by a single actuator. The length of each segment of the DHM was scaled to match the participant's body segment length, while the DHM inertial parameters were scaled based on the participant's height and mass using average anthropometric coefficients.

To retarget the participant's upper-body motion in the Cartesian space, we used a hierarchical quadratic programming (QP) controller based on the OpenSoT library, which computes velocity commands to animate the DHM. The QP objective function consisted of the following tasks and priorities:

- level 1 (top priority tasks): balance (center of mass) position task, feet position task (fixed);
- level 2: Cartesian trajectory tracking of the pelvis and thoracic spine segments (position and orientation), of the right and left shoulder, elbow and wrist (position only), and of the head orientation;

where the reference trajectories for the tracking tasks were the 3D positions and orientations of the Xsens avatar's body segments.

After retargeting the participant's upper-body motion, we used the DHM L5/S1 flexion/extension joint torques estimated with the dynamic simulation to compare the lumbar effort exerted by the participant with and without the exoskeleton. The pipeline for this processing is represented in the following Figure.

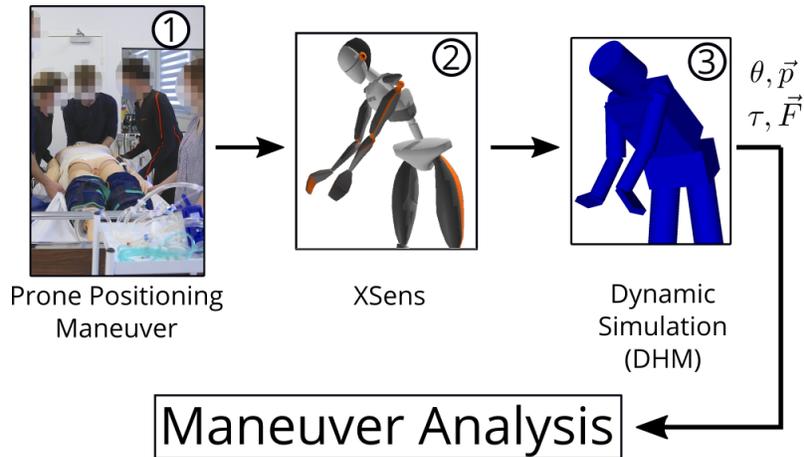

*Figure 6- In the study conducted at the Hospital Simulation Center, the motion of one physician executing the PP maneuver was captured with the Xsens MVN suit (Step 1). We used the whole-body kinematic estimation of the Xsens MVN software (Step 2) as an input to our dynamic simulation with a Digital Human Model (Step 3). The analysis of motion and estimation of human lumbar effort are based on this dynamic simulation.*

## IV. KINEMATIC ANALYSIS

We analyzed the motion of the L5/S1 joint in the sagittal plane during the use of the four exoskeletons in the Prone to Supine (PS) and Supine to Prone (SP) positioning. The following figure displays typical profiles of low-back flexion angle of one participant, for the PS and SP, for all four exoskeletons and without exoskeleton. The joint angle profiles are overall similar for all conditions; variations from one condition to another can be explained by small differences in the manikin's position on the bed and intrinsic variability in the entire maneuver performed by the team.

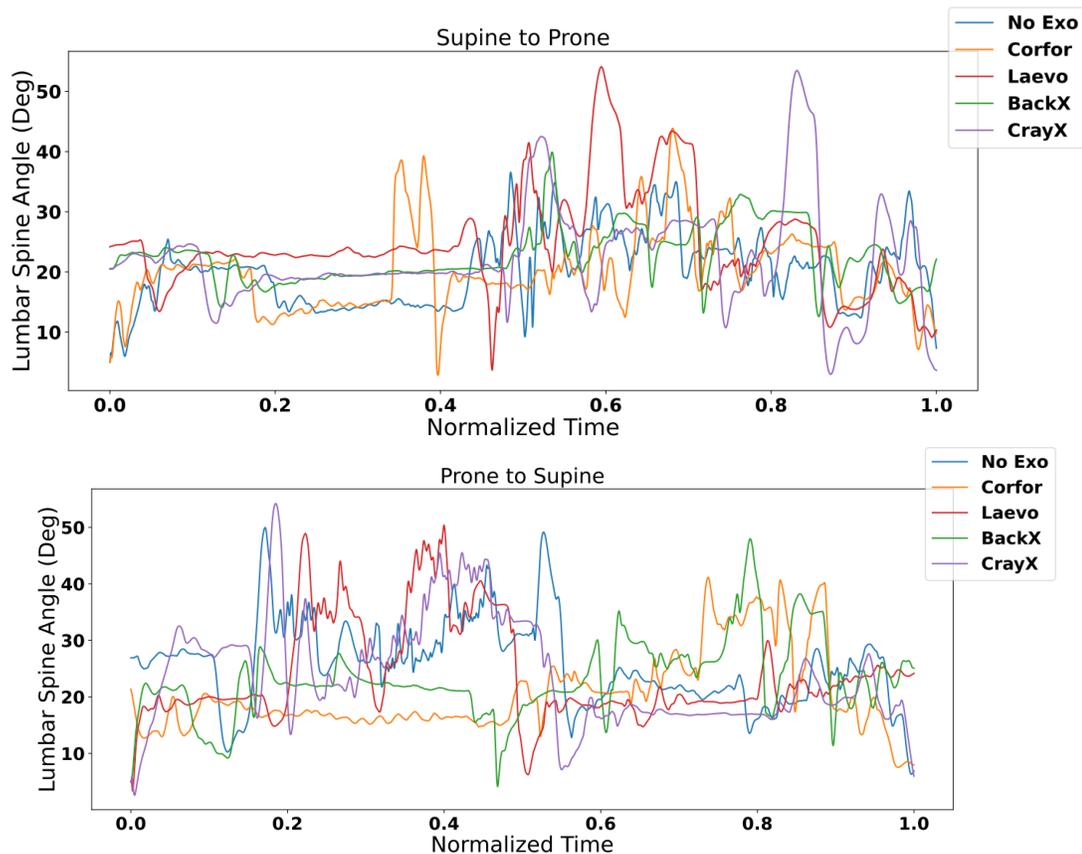

*Figure 7- Lumbar spine flexion angle of one participant performing the PP maneuvers at the Hospital Simulation Center.*

## V. DYNAMIC ANALYSIS

We estimated the human L5/S1 joint torque during the Prone to Supine (PS) and Supine to Prone (SP) positioning, with and without the Laevo exoskeleton, using our DHM simulation.

When the participant is equipped with the exoskeleton, the net torque exerted at the L5/S1 joint to counter the dynamics and gravity effects on the upper-body is a sum of the human-generated torque and of the exoskeleton assistive torque: $\tau_{L5S1} = \tau_{human} + \tau_{exo}$. In order to estimate the human torque, the assistive torque $\tau_{exo}$ provided by each exoskeleton is needed. To compute this torque, one needs the details about the mechatronics design of the platform.

We performed this computation only for the Laevo exoskeleton, since it was the one unanimously perceived by the participants as the most suitable candidate for use during PP maneuvers. Based on the Laevo empirical calibration curve published by Koopman et al. and on the Laevo user manual which specifies that its set of springs provides a maximum torque of 40 Nm, we used the following model to estimate $\tau_{laevo}$:

$$\tau_{laevo} = \begin{cases} k_0 + k_1\theta & \dot{\theta} > 0 \\ k_0 + k_1\theta - k_{loss}, & \dot{\theta} < 0 \end{cases}$$

where $\theta$ is the back flexion angle, $k_0 = -80/3 Nm$ and $k_1 = 4/3 \frac{Nm}{deg}$ are constants that encode the spring linearity in its range of operation from 20 to 50 degrees (with the maximum assistance of 40 Nm at 50 degrees), and $k_{loss} = 10 Nm$ represents frictional losses which introduce hysteresis in the system (numerical values of the model's coefficients were set so that the model matches the calibration curve in [9] as closely as possible).

The following figure displays the estimated joint torque and lumbar flexion angle across time for one PP trial for one participant.

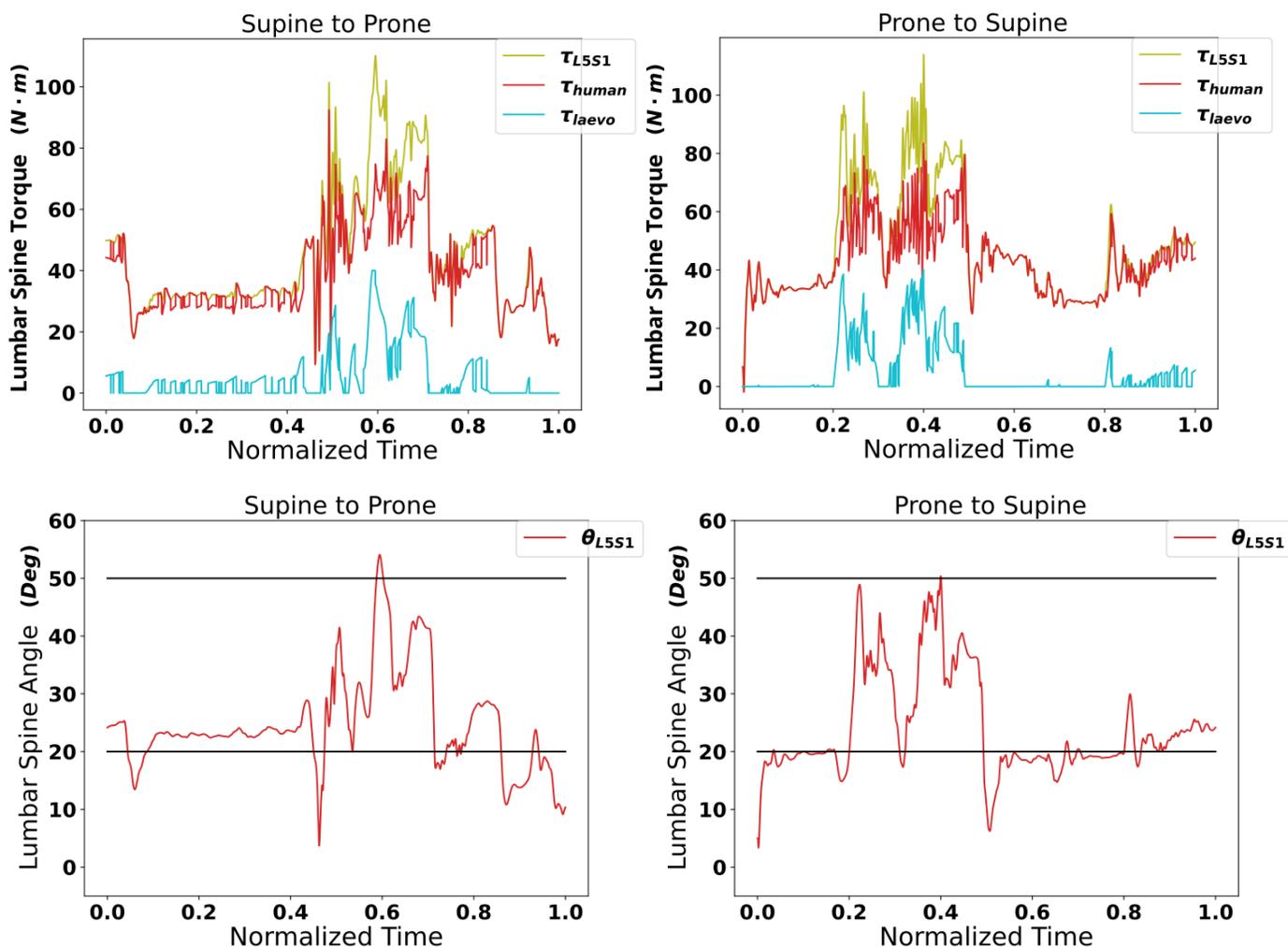

*Figure 8 – Estimation of the human lumbar flexion torque and lumbar flexion angle of one participant performing the PP maneuvers at the Hospital Simulation Center. The torques are estimated with the DHM simulation. The plot reports one PP trial across time.*

## VI. EMG AND ECG PLACEMENT

To study the quantification of the Laevo's assistance, we recorded physiological measures on the participants executing PP maneuvers with and without the Laevo exoskeleton. The participants were equipped with an ECG sensor (Delsys Trigno ECG Biofeedback, 2 channels, bandwidth: 30Hz, ECG sampling rate 4370 sa/sec with onboard Butterworth bandpass filter 40/80 dB/Dec) and 12 surface EMG sensors (Delsys Trigno, EMG sampling rate 4370 sa/sec). The following figure shows the sensors placement on the participant's body, which was done according to Seniam protocol recommendations after abrasion and cleaning with alcohol.

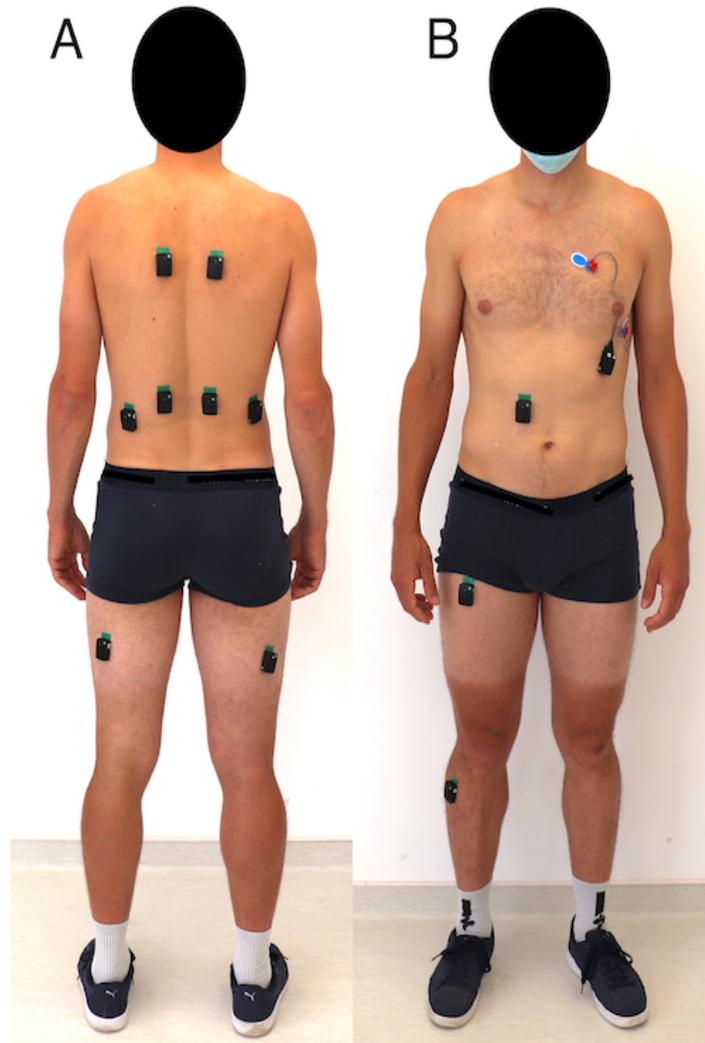

*Figure 9 - ECG and surface EMG placement for the quantification of Laevo's assistance during a PP maneuver. A: EMG sensors on ESL, ESI, TA, BF and GM. B: ECG sensor with 2 electrodes and EMG sensors on RA, RF and TA.*

## VII. DETAILED EMG RESULTS

In the following figure we report the distribution of the EMG signals for 8 right side muscles of one participant, with and without the Laevo exoskeleton.

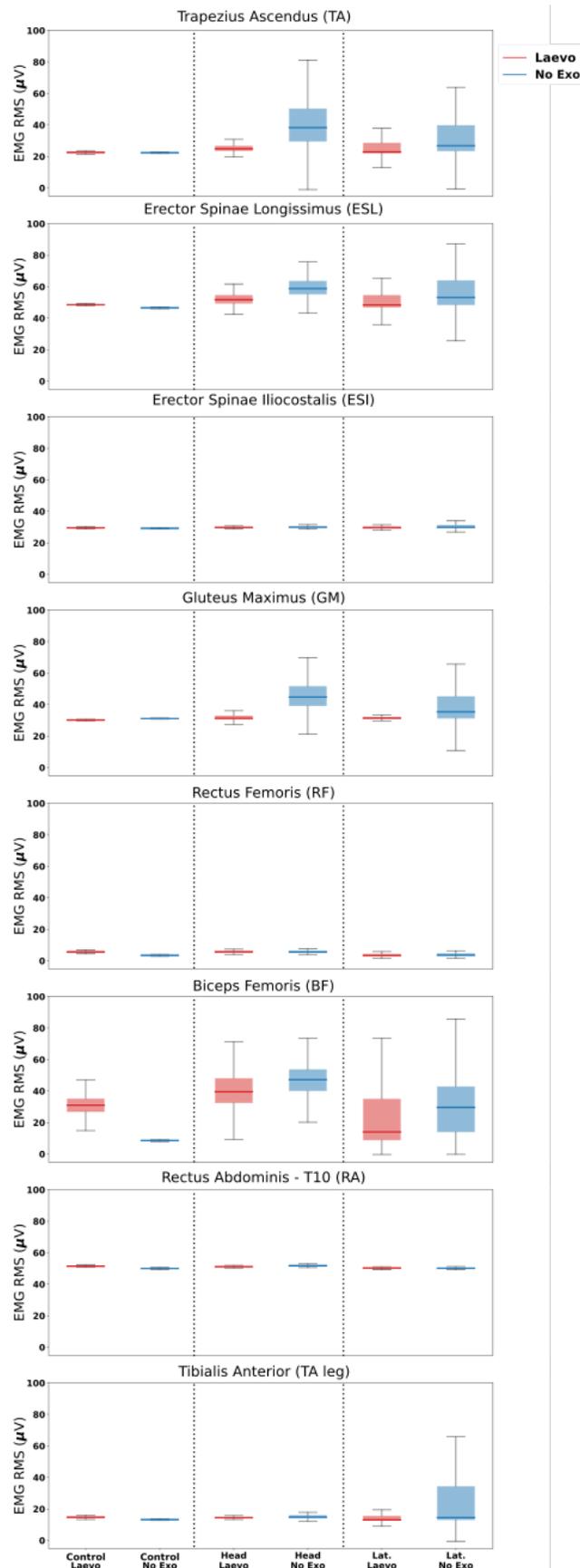

*Figure 10 - Boxplots representing the distribution across time of the EMG signal for 8 right side muscles of one participant with and without the Laevo exoskeleton, during a control condition (i.e., resting) and executing the PP maneuver at the head and at the side of the patient simulator at the Hospital Simulation Center.*

ACKNOWLEDGMENT

The authors wish to thank all the members of the project "ExoTurn" that made it possible to realize this pilot study, as well as their respective institutions for their support. The work was partially funded by the European Commission through the project H2020 AnDy (GA no. 731540).

AUTHORS

**Serena Ivaldi** University of Lorraine, CNRS, Inria, LORIA, F-54000 Nancy, France. Email: serena.ivaldi@inria.fr

**Pauline Maurice** University of Lorraine, CNRS, Inria, LORIA, F-54000 Nancy, France. Email: pauline.maurice@loria.fr

**Waldez Gomes** University of Lorraine, CNRS, Inria, LORIA, F-54000 Nancy, France. Email: waldez.azevedo-gomes-junior@inria.fr

**Jean Theurel** Working Life Department, French National Research and Safety Institute for the Prevention of Occupational Accidents and Diseases (INRS), F-54500 Vandoeuvre-les-Nancy, France. Email: jean.theurel@inrs.fr

**Lien Wioland** Working Life Department, French National Research and Safety Institute for the Prevention of Occupational Accidents and Diseases (INRS), F-54500 Vandoeuvre-les-Nancy, France. Email: lien.wioland@inrs.fr

**Jean-Jacques Atain-Kouadio** Working Life Department, French National Research and Safety Institute for the Prevention of Occupational Accidents and Diseases (INRS), F-54500 Vandoeuvre-les-Nancy, France. Email: jean-jacques.atain-kouadio@inrs.fr

**Laurent Claudon** Working Life Department, French National Research and Safety Institute for the Prevention of Occupational Accidents and Diseases (INRS), F-54500 Vandoeuvre-les-Nancy, France. Email: laurent.claudon@inrs.fr

**Hind Hani** Virtual Hospital of Lorraine, CUESim, University of Lorraine, F-54000 Nancy, France. Email: hind.hani@univ-lorraine.fr

**Antoine Kimmoun** Medical Intensive Care Unit Brabois, Nancy University Hospital, INSERM U1116, University of Lorraine, 54000 Nancy, France. Email: a.kimmoun@chru-nancy.fr

**Jean-Marc Sellal** CHRU-Nancy, Department of Cardiology, F-54000 Nancy, France. Email: jm.sellal@chru-nancy.fr

**Bruno Levy** Medical Intensive Care Unit Brabois, Nancy University Hospital, INSERM U1116, University of Lorraine, 54000 Nancy, France. Email: b.levy@chru-nancy.fr

**Jean Paysant** CHRU-Nancy, Department of Rehabilitation Medicine, EA DevAH, University of Lorraine, F-54000 Nancy, France. Email: jean.paysant@univ-lorraine.fr

**Serguei Malikov** CHRU-Nancy, Inserm 1116, University of Lorraine, F-54000 Nancy, France. Email: s.malikov@chru-nancy.fr

**Bruno Chenuel** CHRU-Nancy, University of Lorraine, University Centre of Sports Medicine and Adapted Physical Activity, Pulmonary Function and Exercise Testing Department, EA DevAH, Department of Medical Physiology, F-54000 Nancy, France. Email: b.chenuel@chru-nancy.fr

**Nicla Settembre** CHRU-Nancy, Inserm 1116, Virtual Hospital of Lorraine, University of Lorraine, F-54000 Nancy, France. Email: nicla.settembre@univ-lorraine.fr